\documentclass[letterpaper, 10 pt, conference]{ieeeconf}  
\IEEEoverridecommandlockouts                              
\usepackage{graphics} 
\usepackage{epsfig} 
\usepackage{times} 
\usepackage{amsmath} 
\usepackage{amssymb}  
\usepackage{algorithm}
\usepackage[noend]{algpseudocode}
\usepackage{xcolor}
\usepackage{balance}
\usepackage{caption}
\usepackage{subcaption} 
\usepackage{cite}
\usepackage{hyperref}

\DeclareMathOperator*{\tr}{tr}
\DeclareMathOperator*{\ts}{ts}
\definecolor{orange}{RGB}{179, 98, 0}

\title{\LARGE \bf
Bayesian Meta-Learning for Few-Shot Policy Adaptation \\Across Robotic Platforms
}

\author{Ali Ghadirzadeh$^{*1}$, Xi Chen$^{*2}$, Petra Poklukar$^2$, Chelsea Finn$^{1}$, M{\aa}rten Bj{\"o}rkman$^2$ and Danica Kragic$^2$
\thanks{
\newline
 $^1$Stanford University, $^2$KTH Royal Institute of Technology \newline
$^*$ denotes equal contribution.}
}

\begin{document}

\maketitle
\thispagestyle{empty}
\pagestyle{empty}

\begin{abstract}
Reinforcement learning methods can achieve significant performance but require a large amount of training data collected on the same robotic platform. A policy trained with expensive data is rendered useless after making even a minor change to the robot hardware. In this paper, we address the challenging problem of adapting a policy, trained to perform a task, to a novel robotic hardware platform given only few demonstrations of robot motion trajectories on the target robot. We formulate it as a few-shot meta-learning problem where the goal is to find a meta-model that captures the common structure shared across different robotic platforms such that \textit{data-efficient} adaptation can be performed. We achieve such adaptation by introducing a learning framework consisting of a probabilistic gradient-based meta-learning algorithm that models the uncertainty arising from the few-shot setting with a low-dimensional latent variable. We experimentally evaluate our framework on a simulated reaching and  a real-robot picking task using 400 simulated robots generated by varying the physical parameters of an existing set of robotic platforms. Our results show that the proposed method can successfully adapt a trained policy to different robotic platforms with novel physical parameters and the superiority of our meta-learning algorithm compared to state-of-the-art methods for the introduced few-shot policy adaptation problem.
\end{abstract}
\section{Introduction}
\label{sec:introduction}

Robot learning methods aim to develop data-driven approaches that can deal with the complexity and the diversity of robotic problems. 
Policy learning algorithms have achieved impressive results but mostly on a fixed robotic hardware; to account for even a minor change in the robot hardware, the policy has to be retrained from scratch.  
Training samples are usually quite expensive in such setups, and in most cases, retraining policies from scratch is not data-efficient. 
This problem becomes more profound when learning complex visuomotor skills which typically require a significant amount of data. 
Therefore, the ability to adapt a policy to changes in the robot hardware using few training samples is an important property of a learning system. 
Likewise, sample efficiency of robot learning algorithms will be significantly improved overall if we can train a policy for a task on a robotic platform once and adapt it to all other similar robots with only few training samples. 
However, training general policies that adapt efficiently to various robots with different morphologies is a challenging problem for existing data-driven methods. Even though generic learning algorithms can train policies to acquire a wide range of manipulation skills \cite{levine2016end, schulman2017proximal, ghadirzadeh2017deep} there is no way to transfer a trained policy to a new robot despite the common structure that is present in different robotic platforms.
\begin{figure}[t]
\centering
\includegraphics[width=0.9\linewidth]{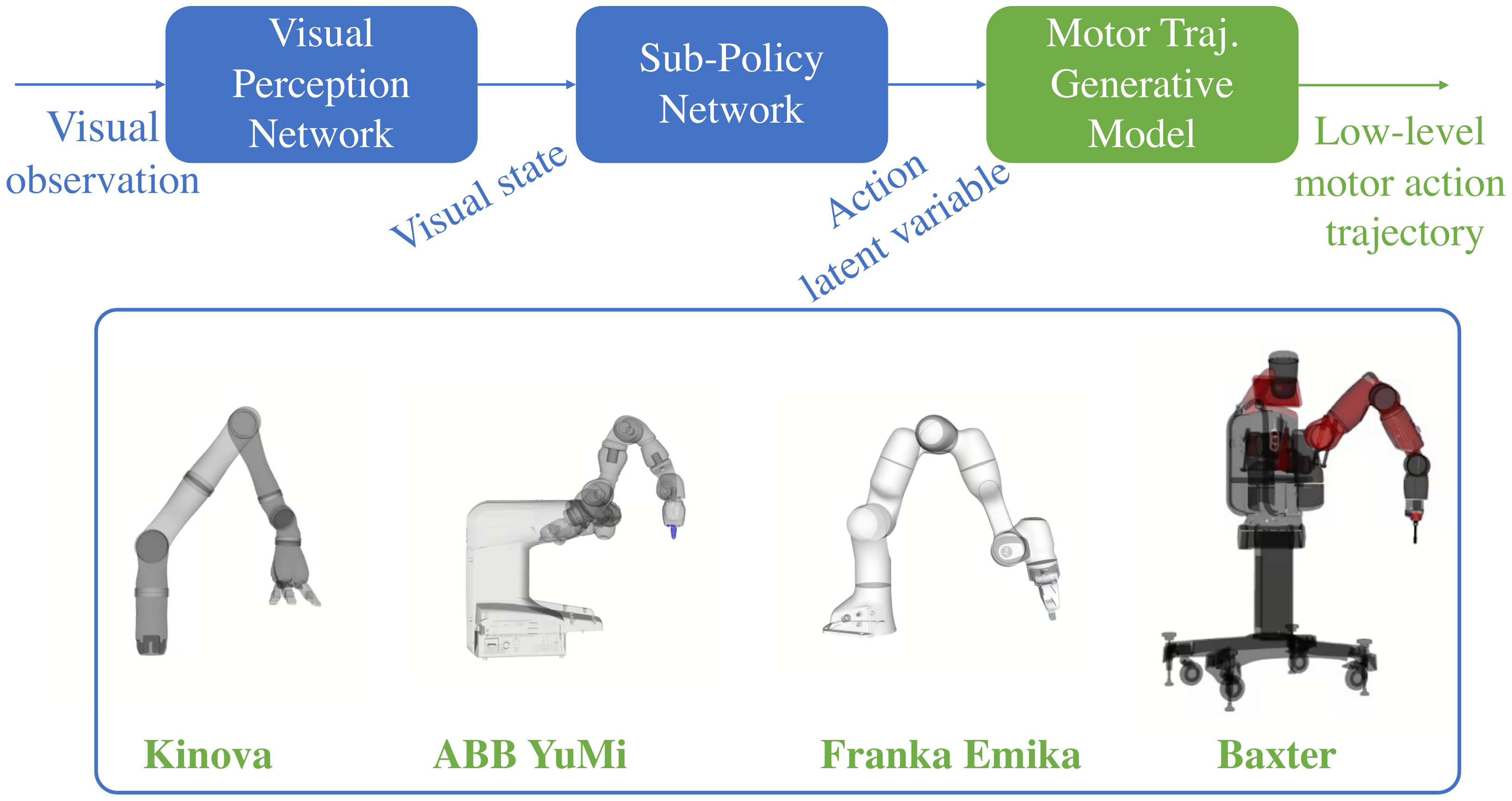}
\caption{Meta-training robot policies on multiple robotic platforms. Top: the block diagram of the policy with the platform independent blocks in blue and the platform dependent block in green. Bottom:  different robotic platforms based on which we generated 400 robots in simulation.}
\label{fig:exp:robots}
\end{figure}

Platform-independent policy training can be formulated as a few-shot learning problem where the goal is to adapt an action-selection policy to a new platform from small amounts of robot data \cite{julian2020efficient, clavera2018learning}.
Such a few-shot learning problem can be addressed by meta-learning \cite{finn2017model} - a learning to learn framework that leverages past knowledge to solve novel tasks more efficiently. In our case, a meta-learning agent can learn a representation of the common structure of joint-level motor commands for a given manipulation task across different robotic platforms.

However, ambiguities arising from the limited
data can make it difficult to find a unique representation of a robotic platform.
Standard meta-learning approaches yield a single solution that is consistent with the given data but might be sub-optimal when the data does not carry sufficient information about the task. 
Probabilistic meta-learning algorithms \cite{finn2018probabilistic, ravi2018amortized, versa_gordon2018metalearning, rusu2019meta} address this problem by leveraging Bayesian inference to generate multiple solutions that are consistent with the few-shot training examples. 
We therefore introduce a probabilistic meta-learning framework that models this uncertainty with a probability distribution from which multiple possible platform representations can be sampled.

The main contributions of this paper are: 
(1) introducing a learning framework to adapt a trained policy to a modified robot hardware or a novel but similar robotic platform, 
(2) introducing a probabilistic meta-learning framework that proposes multiple potential action-selection policies given few demonstrated motion trajectories, and
(3) experimentally evaluating the framework and benchmarking it against some meta-learning algorithms to adapt a trained policy to novel robotic platforms for a simulated  reaching and a real-robot picking task.  
We experimentally evaluate our framework on $400$ robots in simulation which are generated by modifying four existing robot platforms, namely ABB YuMi, Baxter, Franka Emika, and Kinova robots. By training and evaluating a meta-learner on this setup, we show how a trained policy can be adapted to a new robot by providing only a few demonstrated motion trajectories. 
Our experimental results show superior performance compared to the prior meta-learning algorithms \cite{finn2017model, versa_gordon2018metalearning} for a robotic reaching task on different target robots. Also, our real-robot experiments show that the method can successfully adapt a visuomotor policy trained for a grasping task to a YuMi robot with different gripper sizes. 

\section{Related work}
\label{sec:related_work}

\noindent
\textbf{Transfer learning in robotics:}
Few-shot transfer learning has in recent years gained in popularity in robotics research because it addresses both the sample-inefficiency problem of reinforcement learning (RL) algorithms and high costs of recollecting real robot data for every new manipulation task. Prior work focused broadly on the simulation-to-real transfer learning problem \cite{arndt2020meta, arndt2020few ,song2020rapidly,peng2018sim}, few-shot imitation learning \cite{duan2017one,finn2017one,yu2018one,bonardi2020learning,james2018task,butepage2020imitating}, transfer of perception models across different domains \cite{sadeghi2018sim2real, hamalainen2019affordance, chen2020adversarial}, and the transfer of skills across tasks \cite{pinto2017learning, hausman2018learning, gupta2018meta}, dynamics \cite{clavera2018learning, andrychowicz2020learning}, and nonstationary environments \cite{al2018continuous,xie2020deep}. 
However, limited work has been done on transfer learning across robotic platforms  \cite{devin2017learning,dasari2019robonet,huang2020one,schaff2019jointly,chen2018hardware}. 

Devin et al.~\cite{devin2017learning} proposed to decompose a policy network into robot-specific and task-specific modules to facilitate the transfer of knowledge to novel combinations of pre-trained tasks and robots. We similarly decompose a policy into task- and robot-specific components but in contrast to \cite{devin2017learning} our goal is to transfer manipulation skills to a novel robotic platform in a few-shot manner. 
Chen et al. \cite{chen2018hardware} and Schaff et al. \cite{schaff2019jointly} proposed to learn a policy conditioned on an encoding of a robotic hardware, such as the kinematic structure of the robot. Their goal is to learn a general policy that transfers to a new robot either in a zero-shot manner or via fine-tuning \cite{chen2018hardware} and optimizing the robotic hardware together with the policy at the same time \cite{schaff2019jointly}. 
We also learn a robot encoding but instead of feeding it to the policy we use the encoding to initialize a network that is further updated by gradient descent. 
Huang et al. \cite{huang2020one} suggested to train one policy that control each limb of an agent and coordinate between different limbs through a message passing technique. Even though their approach demonstrated appealing results for walking gates, 
it is not clear how to apply it to visuomotor policy training where perception and control are trained together.
Dasari et al. \cite{dasari2019robonet} introduced a large-scale multi-task and multi-robot dynamic model training that can be used with model-based action-selection approaches such as model-predictive control \cite{finn2017deep}. 
We train policies given large-scale datasets constructed in simulation by adjusting physical parameters of different robotic platforms. However, we aim to meta-train an action-selection policy that can then be efficiently adapted to a novel robot given only few motion trajectories from the target robot. 

\noindent
\textbf{Meta-learning:}
Meta-learning is a common approach to solve few-shot learning problems \cite{protonet_NIPS2017_6996, relation_net, taml_8954011, Li2019LGMNetLT}. 
A popular optimization-based  meta-learning method is the model-agnostic meta learning (MAML) framework \cite{finn2017model}. 
However, an important challenge in the few-shot learning setup arises when the few examples do not contain sufficient information to to properly solve a new task \cite{finn2018probabilistic}.
To account for such task ambiguities different probabilistic frameworks based on MAML \cite{finn2018probabilistic, grant2018recasting, bmaml_NIPS2018_7963, rusu2019meta} were introduced as well as a variety of other Bayesian meta learning frameworks \cite{versa_gordon2018metalearning, ravi2018amortized, harrison2018meta, kaddour2020probabilistic, zou2020gradient}.
Similar to \cite{kaddour2020probabilistic, rusu2019meta}, we introduce a low-dimensional meta-task latent variable which we embed into the gradient-based meta-learner following MAML.
In contrast to  \cite{rusu2019meta}, we propose to perform the bi-level gradient descent optimization of MAML directly on the generated network parameters. 
We also find in Section~\ref{sec:experiment} that our approach significantly outperforms VERSA~\cite{versa_gordon2018metalearning} and a method based amortized variational inference (AVI)~\cite{ravi2018amortized}.

\section{Preliminaries}
\label{sec:background}

\noindent
\textbf{Policy training with generative models:} Similar to \cite{ghadirzadeh2020data}, we consider a finite-horizon Markov decision process defined by a tuple $(\mathcal{S}, \mathcal{U}, P, p(s_g), p(r|s_g,\tau))$, where $\mathcal{S}$ denotes both the set of end (terminal) states $s$ and the set of goal states $s_g$, $\mathcal{U}$ is a set of motor actions $u_{m}$ for the robot motor $m$, $P: \mathcal{U} \times \mathcal{S}  \rightarrow \mathbb{R}$ is the state transition probability assuming a fixed initial state, $p(s_g)$ is the goal state distribution and $p(r|s_g,\tau)$ is the probability of the reward $r$ conditioned on a goal state $s_g$ and a fixed-length sequence of open-loop motor actions $\tau = \{u_{t, m}\}$ at time index $t$. 
In order to complete a robotic manipulation task given a goal state $s_g' \sim p(s_g)$, we wish to find a goal-conditioned policy $\pi_\Theta(\tau|s_g')$, parameterized by $\Theta$, that assigns a distribution over the possible sequence of actions $\tau$ that maximize the expected reward $p(r|s_g',\tau)$.
We follow \cite{ghadirzadeh2017deep, ghadirzadeh2020data} and find the parameters $\Theta$ by first introducing an \textit{action latent variable} $\alpha$ and a \textit{trajectory generative model} $p_\vartheta(\tau | \alpha)$, parametrized by $\vartheta$, that maps a latent action sample $\alpha'$ into a sequence of actions $\tau$. Given $\vartheta$, the policy parameters $\Theta = [\theta, \vartheta]$ are then found by marginalizing over $\alpha$ and maximizing the expected reward
\begin{equation}
     \mathcal{J}  =  \max_{\theta}\mathbb{E}_{s_g' \sim p(\cdot), \alpha' \sim \pi_\theta(\cdot|s_g'), \tau' \sim p_\vartheta(\cdot|\alpha')}[p(r | s_g', \tau')],
    \label{eq:rl_objective}
\end{equation}
where a sub-policy $\pi_\theta(\alpha | s_g')$, parametrized by $\theta$,
assigns a distribution over the action latent variable $\alpha$ conditioned on the goal state $s_g'$. Intuitively, the action latent variable $\alpha$ captures the high-level objective of the policy $\pi_\Theta$, while the generative model $p_\vartheta$ translates this objective into a sequence of low-level motor actions.

The complete policy network is illustrated in Figure~\ref{fig:method:meta-test} with blue color.
A goal state $s_g'$ is first mapped into a latent action sample $\alpha'$ by the sub-policy $\pi_\theta$ and then to a sequence of motor actions $\tau'$ by the generative model $p_\vartheta$.
The two parametric models $\pi_\theta$ and $p_\vartheta$ together form the policy $\pi_\Theta$.
The sub-policy $\pi_\theta$ can be trained independently of the robotic hardware platform since the structure of $\alpha$ can be fixed such that the same sub-policy can be used for all robotic platforms without any model adaptation (Sec. \ref{sec:method:policy_training_pi}).
However, the generative model $p_\vartheta$ generates platform-dependent low-level motor actions which requires the parameters $\vartheta$ to be adapted to different platforms. 
The generative model $p_\vartheta$ can be trained on a dataset  
consisting of sequences $\tau$ demonstrated on a given robotic platform
based on variational autoencoders (VAEs) \cite{kingma2013auto, ghadirzadeh2017deep} which maximize the following variational lower bound 
\begin{equation}
     \max_{\upsilon, \vartheta} \mathbb{E}_{\alpha' \sim q_\upsilon(\cdot|\tau)}[\log p_\vartheta(\tau|\alpha')] - D_{KL}(q_\upsilon (\alpha | \tau) || p(\alpha)),
\label{eq:vae}
\end{equation}
where $q_\upsilon(\alpha | \tau)$ denotes the approximate posterior distribution and $p(\alpha)$ the prior over the action latent variable which is set to the standard normal distribution. 

\noindent
\textbf{Multi-task meta learning:} Meta learning can be used to model the common structure shared across different robotic platforms in a way that enables few-shot transfer to related tasks. 
Let a \textit{task} $\mathcal{T}_i$ be defined as the task of generating valid sequence of actions for the $i$th robotic platform.
Let $p(\mathcal{T})$ be an unknown distribution from which we sample infinitely many tasks $\mathcal{T}_i \sim p(\mathcal{T})$ each of which is represented by a dataset $\mathcal{D}_{\mathcal{T}_i}$ consisting of motor action sequences demonstrated on the platform $i$. A common representation of the tasks $\{\mathcal{T}_i\}$ can be learned by training a meta-model using the MAML framework. 
MAML learns a common feature representation or network initialization to achieve an optimal solution on a novel task with only a small number of gradient steps. 
MAML consists of (a) a \textit{meta-train phase} in which the 
meta-model is trained, and (b) \textit{adaptation or meta-test phase} in which the meta-model is adapted to a novel task using gradient descent. The meta-model $\phi$ in phase (a) is trained by optimizing the following objective 
\begin{equation}
    \min_{\phi} \sum_{\mathcal{T}_i \sim p(\mathcal{T})} \mathcal{L}(\phi - \lambda \nabla_\phi \mathcal{L} (\phi, \mathcal{D}_{\mathcal{T}_i}^{\tr}), \mathcal{D}_{\mathcal{T}_i}^{\ts}),
\end{equation}
where, $\lambda$ is the learning rate which itself is a trainable parameter, and $\mathcal{D}_{\mathcal{T}_i}^{\tr}$ and $\mathcal{D}_{\mathcal{T}_i}^{\ts}$ are the support and the  query sets, respectively, such that their union forms the entire task $\mathcal{T}_i$  dataset $\mathcal{D}_{i} = \mathcal{D}_{\mathcal{T}_i}^{\tr} \sqcup \mathcal{D}_{\mathcal{T}_i}^{\ts}$.
Meta-learning methods optimize for few-shot generalization in a nested scheme consisting of an inner loop, in which the model is adapted to individual tasks using the support set $\mathcal{D}_{\mathcal{T}_i}^{\tr}$, and an outer loop, in which  the meta-model's parameters are updated by optimizing the sum of the task-specific losses on the query sets $\mathcal{D}_{\mathcal{T}_i}^{\ts}$.

\section{Few-shot policy training based on meta-learning}
\label{sec:method}

In this section, we introduce our probabilistic meta-learning algorithm that outputs a probability distribution over the meta-model parameters. Using this distribution, we sample the parameters of the trajectory generative models that are used to model a given set of few demonstrated trajectories. We formally define the problem set-up in Sec. \ref{sec:method:problem_setting}, and introduce the framework in Sec. \ref{sec:method:meta-learning}. Finally, we explain the training of the sub-policy $\pi_\theta(\alpha | s_g)$ in Sec.~\ref{sec:method:policy_training_pi}.

\subsection{Platform-independent policy training setup} 
\label{sec:method:problem_setting} 
\begin{figure}[h]
\centering
\includegraphics[width=1.0\linewidth]{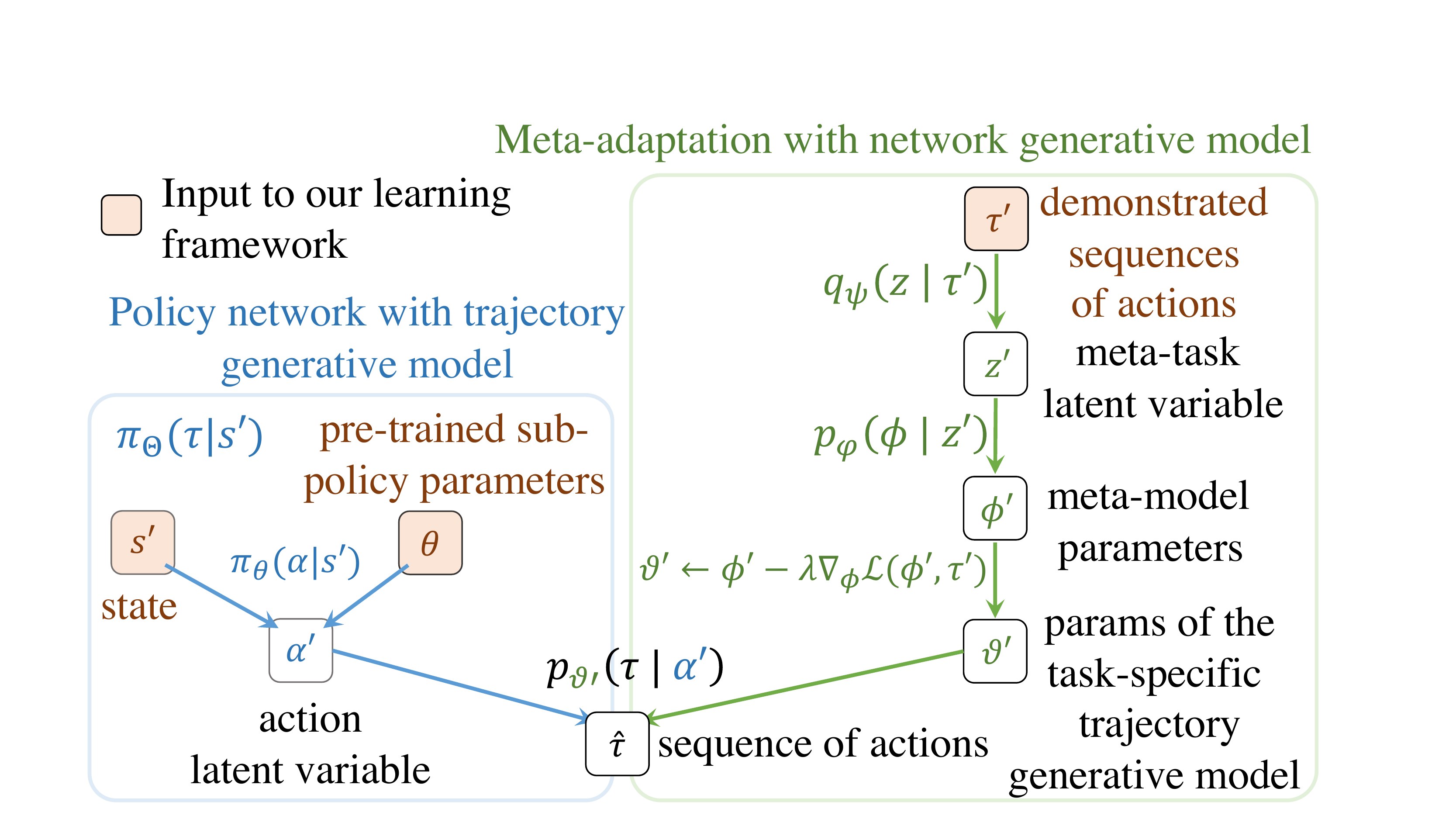}
\caption{The computation graph of the action-selection policy at meta-test time.}
\label{fig:method:meta-test}
\end{figure}

Our goal is to devise a meta-learning algorithm that adapts the policy $\pi_\Theta$, more specifically, the trajectory generative model $p_\vartheta(\tau|\alpha)$, shown in Figure~\ref{fig:method:meta-test}, to an unseen robotic platform, i.e., a novel meta-task $\mathcal{T}_{\text{novel}} \sim p(\mathcal{T})$ in a few-shot manner given the set of support sequence of actions $\mathcal{D}^{\tr}_{\mathcal{T}_{\text{novel}}}$. 
The meta-model is trained on  \textit{meta-train datasets} $\mathcal{D}_{i}$  
containing a valid sequence of motor actions to run on the $i$th robot. 
The datasets are divided into support and query sets containing support sequence of actions $\tau^{\tr}_i \in \mathcal{D}^{\tr}_{\mathcal{T}_i}$ and query sequence of actions $\tau^{\ts}_i \in \mathcal{D}^{\ts}_{\mathcal{T}_i}$.  

The platforms are generated in simulation by choosing a robot model and randomly adjusting the length of different robot links. Using existing motion planners, such as rapidly-exploring random tree (RRT) \cite{karaman2010incremental}, we then generate several sequences of motor actions $\tau$ that complete the desired goal in simulation. The generated trajectories $\tau$ and the label of the robotic platform $i$ form one meta-task dataset $\mathcal{D}_{i}$, while their union over all meta-tasks forms the meta-train dataset $\{\mathcal{D}_{i}\}$. 
\subsection{Our proposed probabilistic meta-learning framework}
\label{sec:method:meta-learning}
Since the small size of a dataset representing a new meta-task gives rise to ambiguities in the meta-model parameters, we extend MAML to a probabilistic setting.
The idea of our probabilistic meta-learning algorithm is to represent meta-tasks in a low-dimensional latent space. Similar to \cite{rusu2019meta}, we obtain such low dimensional representation of the meta-tasks
by introducing a \textit{meta-task latent variable} $z$ and a \textit{network generative model} $p_\varphi(\phi |z)$, parametrized by $\varphi$, that generates the high dimensional meta-parameters $\phi$ conditioned on the representation $z$. Similar to the original MAML's inner loop optimization, we perform gradient descent optimization on the meta-parameters $\phi$, 
as opposed to \cite{rusu2019meta} that performs the gradient descent in the task latent space. 
Our choice is a more straightforward extension of MAML into a conditional and probabilistic meta-learning algorithm while still outperforming state-of-the-art meta-learning algorithms on this problem setting. We leave the comparison between our meta-learning algorithm and \cite{rusu2019meta} to our future work. 

Using the meta-task latent variable $z$ we can first generate different initial meta-models that are well-suited for the few-shot gradient descent updates and then choose the one that results in the best performance on the given novel platform. 
The variable $z$ is sampled from a \textit{variational distribution} $q_\psi(z | \tau)$, parametrized by $\psi$, which is trained jointly with the network generative model $p_\varphi$ using the meta-train dataset $\{\mathcal{D}_{i}\}$ as explained next. Note that we choose $q_\psi$ to be a Gaussian distribution.

During the meta-train phase (a) the goal is to find the parameters of the variational distribution $q_\psi$ and network generative model $p_\varphi$, that yield a distribution over the possible meta-model parameters $\phi$. By sampling from this distribution, different platform-dependent trajectory generative models $p_\vartheta$ are generated that can be efficiently optimized similar to MAML by one or few gradient updates.  
More precisely, $\psi$ and $\varphi$ are optimized by maximizing the following variational lower bound \cite{kingma2013auto}
\begin{equation}
\begin{split}
    \max_{\varphi,\psi} \sum_{i} \mathbb{E}_{ z_i \sim q_\psi(\cdot|\tau^{\tr}_i), \phi_i \sim p_\varphi(\cdot | z_i)  } [\log p_{\vartheta_i}(\tau^{\ts}_i)] \\
    - \beta D_{KL}(q_\psi(z|\tau^{\tr}_i) || p(z)), 
    \label{eq:our_method_loss}
\end{split}
\end{equation}
where, $p(z)$ is the prior over the meta-task latent variable $z$ and $\vartheta_i = \phi_i - \lambda \nabla_{\phi} \mathcal{L}(\phi_i, \mathcal{D}^{\tr}_{\mathcal{T}_i})$ is the gradient step as in the MAML adaptation step and $\beta$ is a scalar parameter that balances the two loss terms. 
Note that we omitted the action latent variable $\alpha$ from the notation $p_{\vartheta_i}(\tau | \alpha)$ in Eq.~\eqref{eq:our_method_loss} for the sake of clarity. Please refer to Sec.~\ref{sec:method:policy_training_pi} for more details. 

During the meta-test phase (b), illustrated in Fig.~\ref{fig:method:meta-test} with green color, a few sequences of motor actions from a new robotic platform are given to the algorithm which then returns a distribution over the parameters of the
trajectory generative model $\vartheta$. More specifically, some sequences of motor actions  $\tau' \in \mathcal{D}^{\tr}_{\mathcal{T}_{\text{novel}}}$ are first recorded given human demonstrations. The action sequences are then given to the variational distribution $q_\psi$ from which we sample several different latent task representations $\{z_j \sim q_\psi(\cdot | \tau')\}$. These are further mapped to the meta-parameters $\{\phi_j\}$ by the network generative model $p_\varphi$. Finally, a set of task parameters $\{\vartheta_j\}$ is found by applying the gradient descent rule on the loss functions $\{\mathcal{L}(\phi_j, \mathcal{D}^{\tr}_{\mathcal{T}_{\text{novel}}})\}$.

\subsection{Training the sub-policy} 
\label{sec:method:policy_training_pi}
In this section, we explain how we can train a \textit{single sub-policy} $\pi_\theta(\alpha | s_g)$ that can be used with different trajectory generative models on different robotic platforms. 
The sub-policy is trained on only one robot and the training is performed independently and prior to the training the meta-model in phase (a). 

The training of the sub-policy $\pi_\theta(\alpha | s_g)$ as described in \cite{ghadirzadeh2020data} requires a trained generative model $p_\vartheta(\tau | \alpha)$ which in turn requires a training dataset of the motor trajectories. We therefore choose one robotic platform $i_*$ for which we can obtain a training dataset of motor action sequences $\mathcal{D}_{\mathcal{T}_{i_*}}$. We train a VAE on $\mathcal{D}_{\mathcal{T}_{i_*}}$ by optimizing Eq.~\eqref{eq:vae}. The obtained generative model $p_{\vartheta_{i_*}}$ is only used to train the sub-policy $\pi_\theta(\alpha | s_g)$ with the Expectation-Maximization algorithm introduced in \cite{ghadirzadeh2020data}.

However, to be able to use the obtained sub-policy with different trajectory generative models on different robotic platforms, we need to ensure a consistent structure of the action latent variable $\alpha$. We define a set of trajectories $\{\tau'_i\}$ to be \textit{consistent}, if they result in the same end state $s$ after executing them on the respective robotic platform. For a set of consistent trajectories, the goal is then to obtain the same latent action sample $\alpha'$ for all robotic platforms.
This is achieved using $\mathcal{D}_{\mathcal{T}_{i_*}}$ and the previously trained VAE encoder $q_{\upsilon_{i_*}}$  introduced in Eq.~\ref{eq:vae}. In particular, a given trajectory $\tau'$, regardless the robotic platform, is matched with a consistent trajectory $\tau'_* \in \mathcal{D}_{\mathcal{T}_{i_*}}$ by comparing the end states. The latter is mapped to a latent action sample $\alpha' \sim q_{\upsilon_{i_*}}(\alpha | \tau'_*)$. The resulting $\alpha'$ and considered trajectory $\tau'$ can then be used to train the meta-model in Eq.~\eqref{eq:our_method_loss}.

\section{Experiments}
\label{sec:experiment}
We apply our proposed learning framework to a robotic reaching task in which the end-effector  of the robot has to reach to different 3D target positions, and a visuomotor picking task similar to \cite{chen2020adversarial} in which a YuMi robot is trained to pick different objects in cluttered background given image pixels as the input. We experimentally evaluate 
(1) the generality of our learned meta-model by training it on three different scenarios described below,
(2) the benefits of the probabilistic formulation of our method by benchmarking it to MAML, and (3) the benefits of gradient based optimization by benchmarking it to two gradient-free methods described in Sec.~\ref{sec:experiment:benchmark}.
We experimentally evaluate the generality of our meta-learning method when adapting the policy to (i) a slightly modified robotic hardware generated by changing the length of a link of the robot, and to (ii) a completely new robotic platform which is not used during the meta-training phase. 
We created $400$ different robots in simulation by adjusting the mechanical parameters of four 7 degree-of-freedom robotic platforms, ABB YuMi, Kinova, Franka Emika and Baxter, shown in Figure~\ref{fig:exp:robots}.
We consider three different scenarios: 
(a) meta-training given data from robots built on one of the platforms (100 robots), and adapting to  novel robots built on the same platform,
(b) meta-training given all of the $400$ simulated robots and adapting as in (a), and
(c) meta-training given data from robots built on only three platforms (300 robots) and adapting to novel robots built on the excluded platform. The scenarios (a) and (b) are used to evaluate criteria (i), while scenario (c) is used to evaluate criteria (ii).

At meta-test time, we sample $n = 3$ novel robots for each platform introduced in Figure~\ref{fig:exp:robots}. We used 5 demonstrated motion trajectories to adapt the trajectory generative model $p_{\vartheta_n}$, produced by the meta-learning agent, to the platform $n$. The demonstrated trajectories are consistent across platforms as defined in Sec.~\ref{sec:method:policy_training_pi}.
We evaluate the performance of the obtained policy $\pi_\Theta$ on the novel robot $n$ using Eq.~\eqref{eq:rl_objective}. 

\subsection{Network architectures} 
\label{sec:experiment:arch}
In this section, we introduce the network architectures of the sub-policy, the trajectory generative model, and our proposed meta-learner model.
Following \cite{chen2020adversarial}, the sub-policy $\pi_\theta(\alpha | s)$ consists of four convolutional layers with $[64,32,32,1]$ channels and $[5\times5]$ filters followed by three fully connected dense layers with $[1058\times16, 16\times16, 16\times6]$ neurons in each layer. 
For the picking task, the sub-policy maps an image of the size $60\times105$, and for the reaching task a 2D target position into a 6D action latent variable $\alpha$. In the latter case, the convolutional layers are excluded. 
The generative model $p_\vartheta(\tau | \alpha)$, consisting of two dense layers with $[6\times8,8\times98]$ neurons, then maps the 6D action latent variable into a trajectory of motor actions of the size $7\times14$, i.e., $7$ motor actions over $14$ time-steps. The motor actions are joint level position commands sent to a position controller. 

Our proposed meta-model consists of an encoder $q_\psi(z|\tau)$ and a network generative model $p_\varphi(\phi|z)$. 
The encoder $q_\psi$ maps 5 demonstrated motor trajectories of the size $7\times14$ into the 2D task latent variable $z$ using 4 dense layers with $[490\times128,128\times64, 64\times32, 32\times2]$ neurons, respectively. 
The network generative model $p_\varphi$ consists of 5 dense layers of size $[2\times32, 32\times64,64\times128, 128\times256, 256\times938]$. 
It generates the parameters of the meta-model neural network (having 832 weights and 106 biases) given the 2D task latent variable $z$ as the input. 
The meta-model has the same architecture as the generative model, and is adapted to a platform-specific generative model after one or few gradient updates.
The Rectified Linear Unit (ReLU) is used as the non-linearity in our network architecture. 

\begin{figure}[h]
\centering
\includegraphics[width=0.9\linewidth]{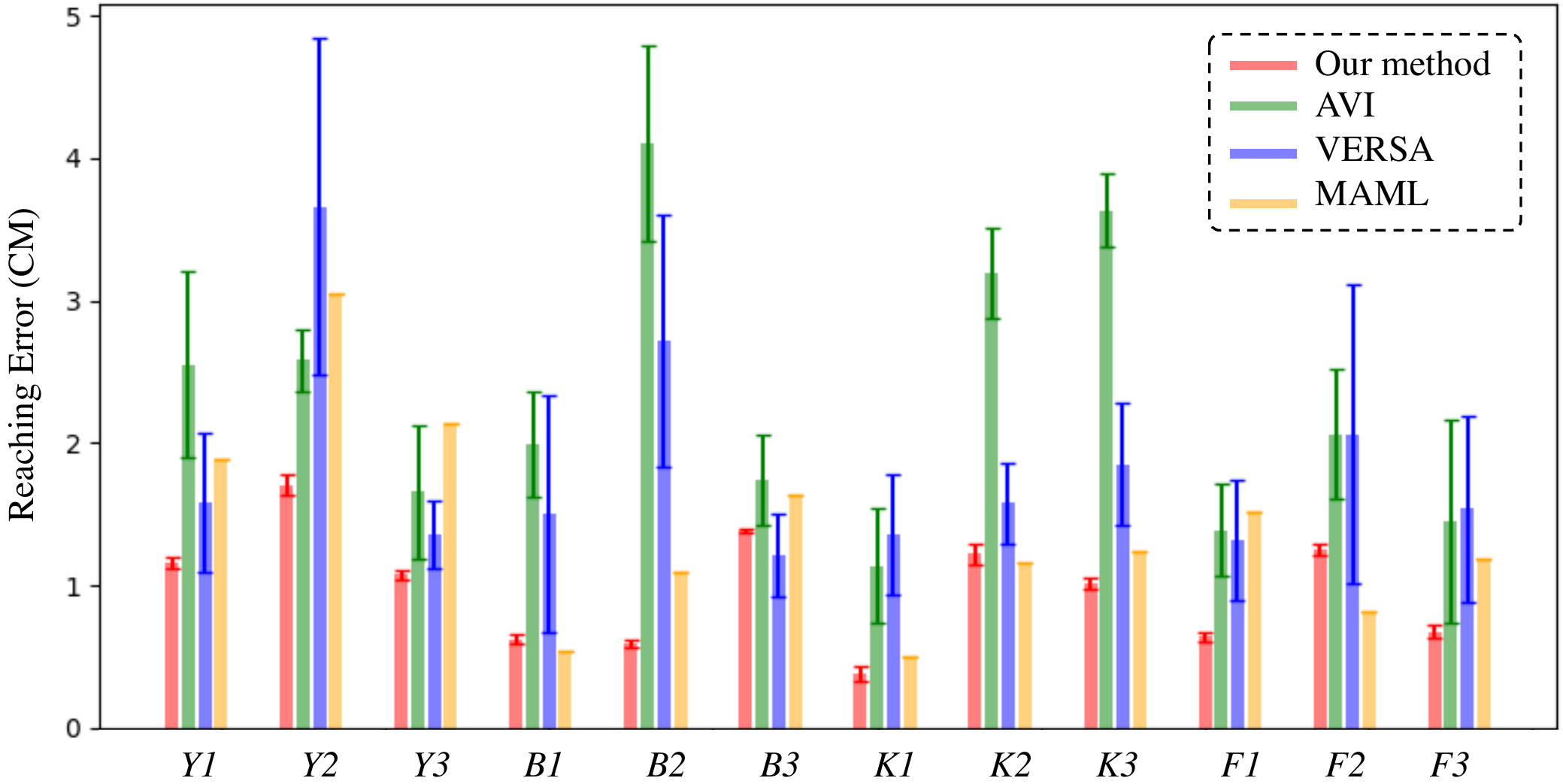}
\caption{Scenario (a): The average and the 95\% confidence interval of the reaching error (in cm) of the adapted policies for all methods. The meta-learner is trained and tested on the data generated by the same platform. \emph{Y} stands for YuMi, \emph{B} for Baxter, \emph{K} for Kinova, and \emph{F} for Franka Emika, while the numbers denote the index of the test robot. }
\label{fig:exp:scenario1}
\vspace{-0.35cm}
\end{figure}

\subsection{Benchmarking methods} 
\label{sec:experiment:benchmark}
We benchmark our method with MAML \cite{finn2017model}, VERSA \cite{versa_gordon2018metalearning} and a method based on amortized variational inferences (AVI) \cite{ravi2018amortized} using three different meta-train and meta-test datasets constructed according to the three scenarios. 
These methods are used to adapt the parameters of the trajectory generative model 
given few demonstrated motion trajectories during the meta-test phase. 

\noindent
\textbf{MAML}: The meta-model in MAML is obtained with
the same architecture as the trajectory generative model (introduced in Sec.~\ref{sec:experiment:arch}) and is adapted to a platform-specific generative model using one gradient update. 

\noindent
\textbf{VERSA}:
We adopt the VERSA model used for few-shot view reconstruction in \cite{versa_gordon2018metalearning} to obtain the motor trajectory generative model. 
The meta-learner consists of an encoder $q_\psi(z|\tau)$ that similarly to our approach maps 5 motor trajectories into a 2D task latent variable. 
However, the task latent variable is directly given as an input to the trajectory generative model as opposed to the network generative models as in our method. Therefore, in VERSA, the parameters of the trajectory generative model are  \textit{platform independent}.

\noindent
\textbf{AVI}:
In order to study the benefit of the gradient descent component of our method, we exclude the gradient optimization step from our method.
The meta-learner in this case consists of a task encoder and a network generative model.

For both VERSA and AVI, we use the same network architecture as described in Sec.~\ref{sec:experiment:arch}. We train all the methods, including ours, for 1000 epochs 
using ADAM optimizer~\cite{kingma2014adam} with learning rate $1e^{-4}$. For our method we use $\beta=5e^{-3}$. All hyper-parameters are found by searching for optimal values. 

\begin{figure}[h]
\centering
\includegraphics[width=0.9\linewidth]{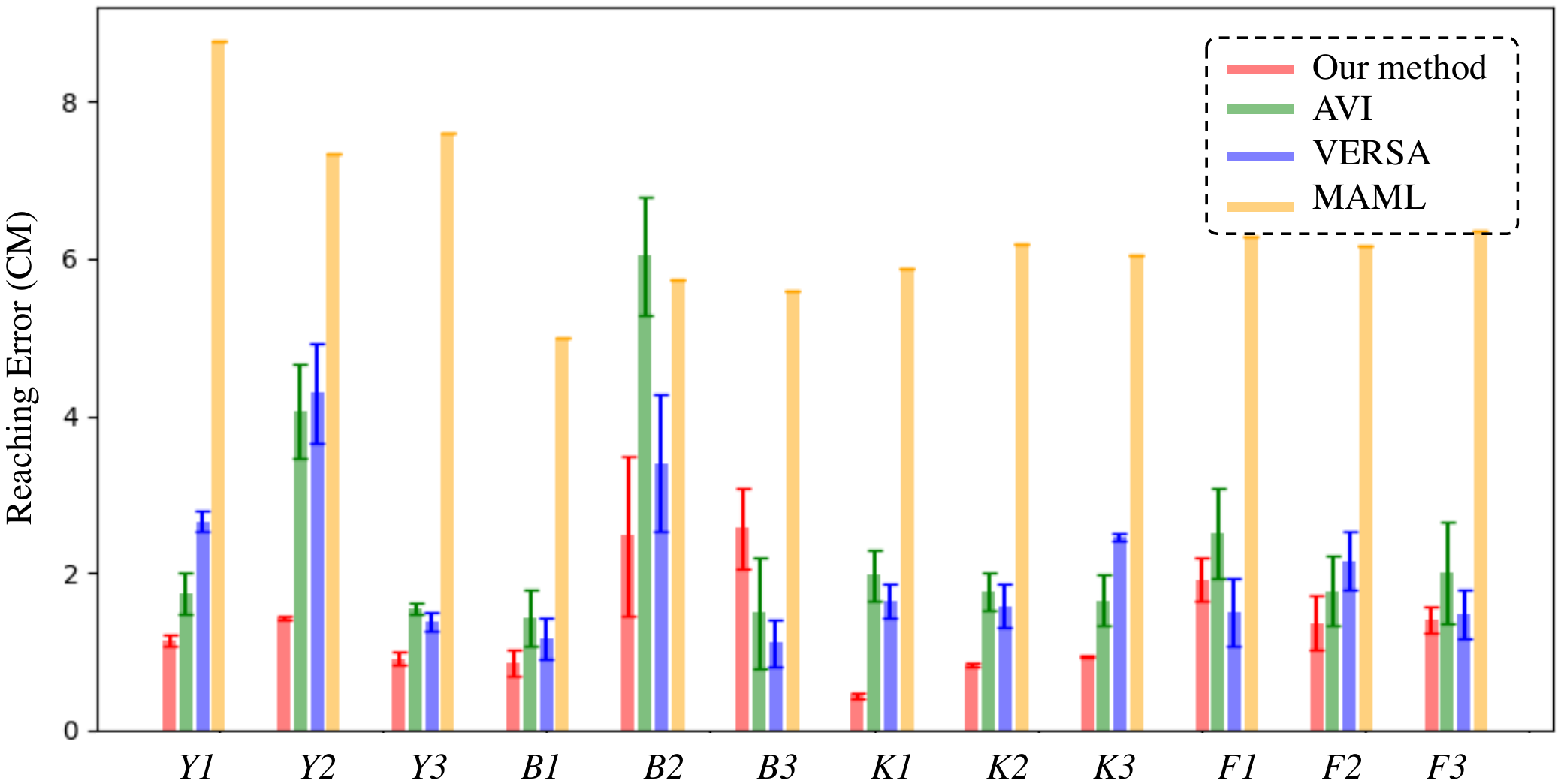}
\caption{Scenario (b): The average and 95\% confidence interval of the reaching error as in Fig.\ref{fig:exp:scenario1}. The meta-learner is trained and tested on the data generated by all four platforms.}
\label{fig:exp:scenario2}
\vspace{-0.3cm}
\end{figure}
\begin{figure}[h]
\centering
\includegraphics[width=0.9\linewidth]{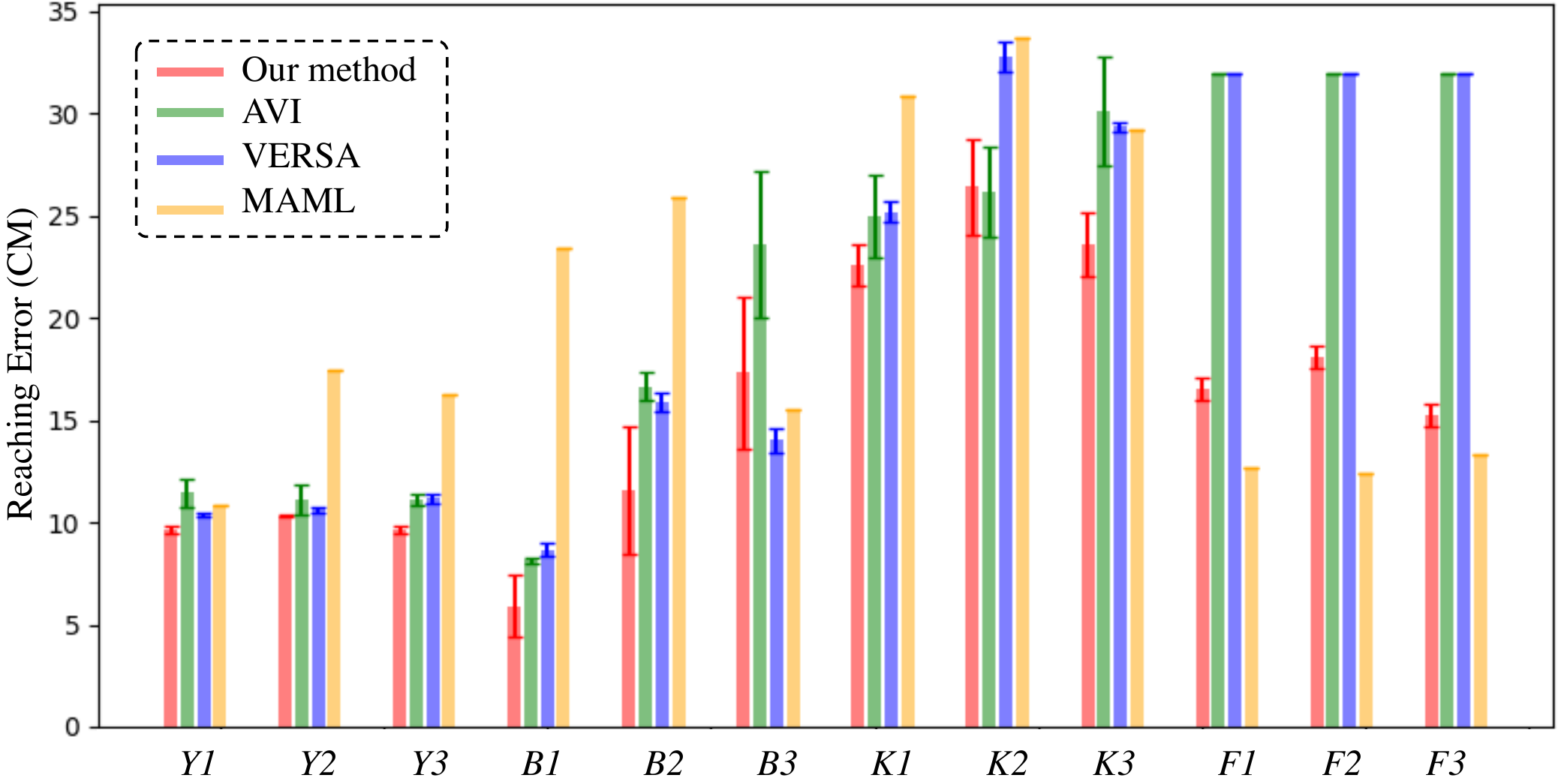}
\caption{Scenario (c): The average and 95\% confidence interval of the reaching error as in Fig.\ref{fig:exp:scenario1}. The meta-learner is trained on the data from all platforms except the one that is used for meta-testing.}
\label{fig:exp:scenario3}
\end{figure}

\subsection{Evaluating the meta-learning agent} 
\label{sec:experiment:meta}
For our method, VERSA and AVI, we sampled $j = 20$ policies from the learned distribution based on the 5 motion trajectories. In Fig.~\ref{fig:exp:scenario1}-\ref{fig:exp:scenario3} we provide the average and $95\%$ confidence interval of the reaching error on the obtained 20 polices 
for the robotic reaching task. For MAML we report the reaching error of the obtained single solution. Our experimental results find that our method outperforms MAML, VERSA, and AVI in most cases for all of the three scenarios.
In scenario (a), in which the meta-train and meta-test datasets include data from only one platform, we observe that in most cases our method performs considerably better than the others, also with lower variance (Fig.~\ref{fig:exp:scenario1}). 
This experiment suggests that our few-shot learning approach performs well when adapting the policy to changes in the \textit{same} robot hardware.

For scenario (b), in which the meta-train dataset contains data from all 4 robot platforms, MAML performs significantly worse compared to scenario (a) (Fig~\ref{fig:exp:scenario2}). 
This means that providing a more diverse dataset deteriorates MAML adaptation performance. Such diversity makes it difficult for MAML to find a network initialization from which different meta-tasks can be solved. 
For this scenario, our method provides the best average performance in most cases, and achieves significant gains compared to prior methods on the YuMi and Kinova platforms. 
These results demonstrate the effectiveness of adaptation to a wider range of meta-tasks. 

\begin{figure}[h]
    \centering
  \includegraphics[width=0.75\linewidth]{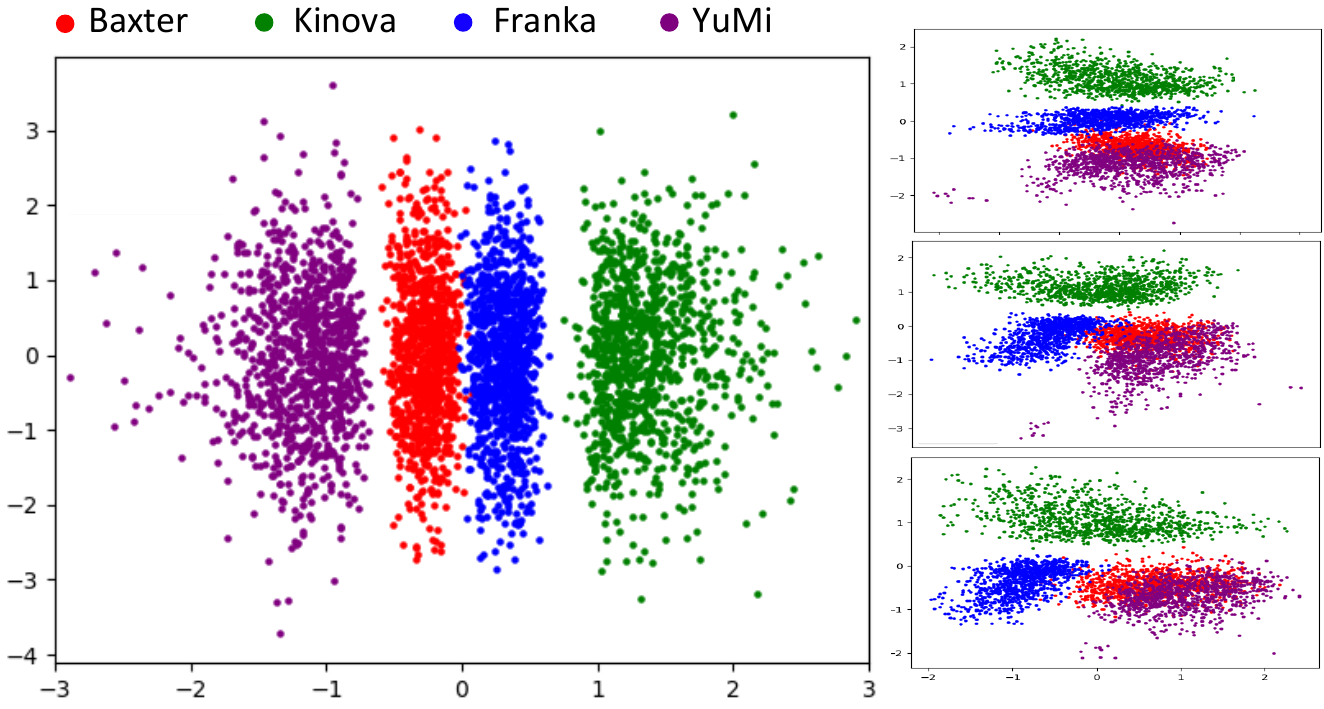}
  \caption{Visualization of the meta-task latent space. Left: meta-task latent samples corresponding to the training dataset of a model trained for scenario (b). Right: meta-task latent samples corresponding to the training dataset of a model trained for scenario (c) as well as to the testing dataset of the excluded robot (in red). }
  \label{fig:exp:task_latent_var}
  \vspace{-0.35cm}
\end{figure}
Next, we evaluate our method based on criteria (ii) using scenario (c), in which the meta-learner should adapt to out-of-distribution tasks (Fig.~\ref{fig:exp:scenario3}). We observe that the methods do not yield satisfactory performance. The results of our method are slightly better for the Baxter and YuMi robots, potentially due to the similarities between these two robots. 
To further analyze the results, we also studied the latent structure of our method for models trained in scenarios (b) and (c). 
In Fig.~\ref{fig:exp:task_latent_var} we illustrate the mean values of the meta-task latent variable $z$ obtained from the Gaussian encoder $q_\psi$ corresponding to all 
the robots used to generate the meta-train datasets in scenarios (b) and (c). 
The left figure illustrates the latent space of a model trained on scenario (b). 
We see that our method well separates the four robotic platforms in the meta-task latent space.  
The right three figures in Fig.~\ref{fig:exp:task_latent_var} illustrate the meta-task latent space of three independently trained models for scenario (c) in which Baxter robots are excluded from the meta-train phase. The figures also show the Baxter robots encoded at meta-test time.
The results suggest that the Baxter robots (at meta-test) look similarly to the YuMi robots that were used to train the meta-learner, which is consistent with the results visualised in Fig.~\ref{fig:exp:scenario3}.

\begin{figure}[h]
\centering
\includegraphics[width=0.8\linewidth]{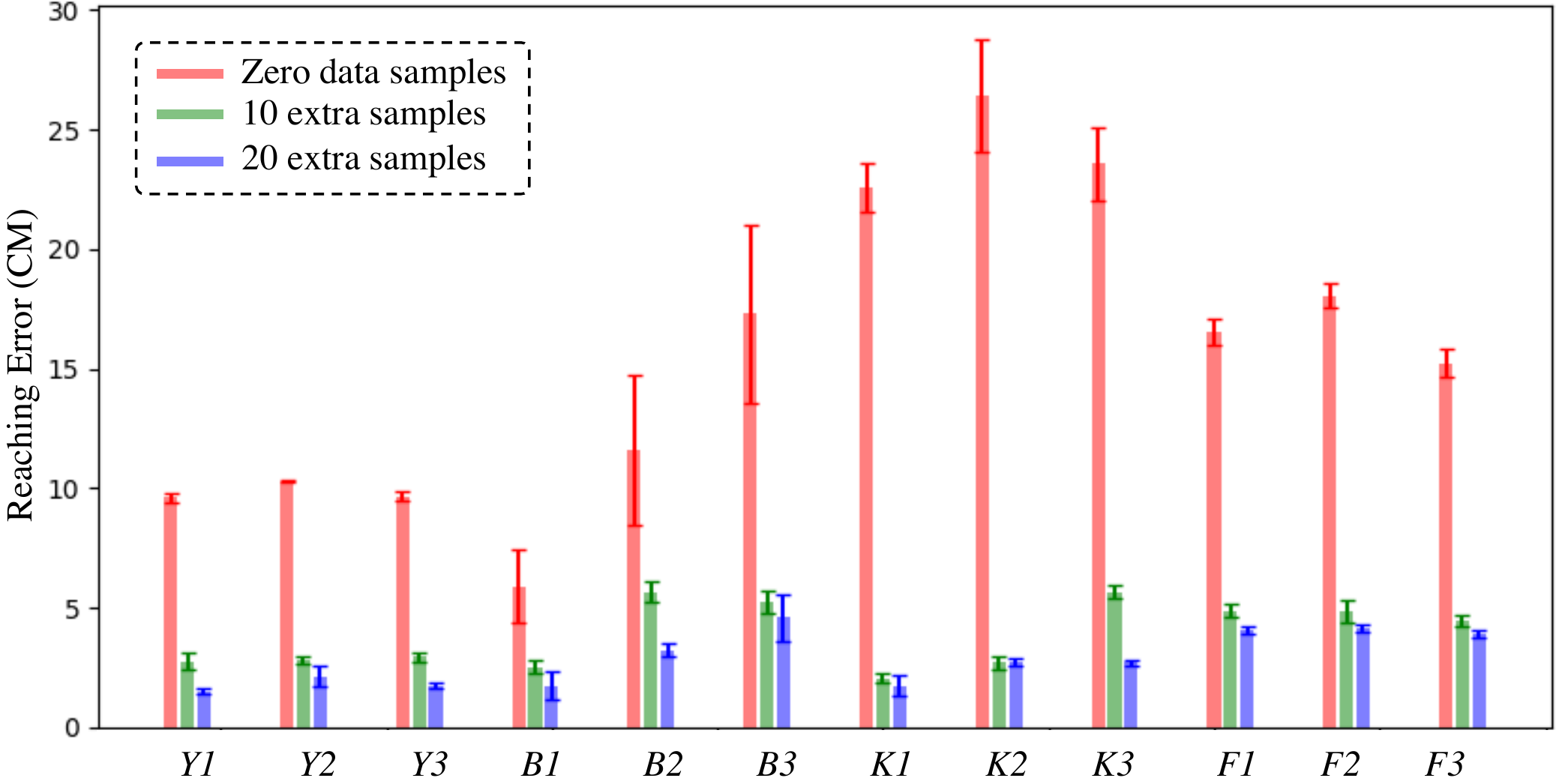}
\caption{The average and 95\% confidence interval of the reaching error as in Fig.\ref{fig:exp:scenario1}. Our meta-learner is trained on the data from all platforms except the one used for meta-testing of which 0, 10 and 20 meta-task samples are used during meta-training.}
\label{fig:exp:scenario4}
\vspace{-0.4cm}
\end{figure}

Beyond these visualizations, we add one new experiment to understand how we might mitigate the challenge of adapting to out-of-distribution tasks. In particular, to avoid meta-testing on completely out-of-distribution tasks as in scenario (c), we included $10$ and $20$ meta-task samples of the excluded platform in the meta-train dataset and repeated the experiment. The results (Fig.~\ref{fig:exp:scenario4}) are significantly better than the results reported for scenario (c), suggesting that a relatively small amount of meta-training data on these platforms is particularly helpful for alleviating out-of-distribution challenges. 

Finally, we present our experimental results for the visuomotor picking task on a real YuMi robot. First, we built 4 different robot grippers that are $-1$ CM, $0$ CM, $+1$ CM and $+2$ CM longer than the original YuMi robot gripper. Using our method, we meta-trained given 100 simulated robots as described in scenario (a) and meta-tested on the real YuMi robot with the four different grippers. On average, we obtained  $90.8\%$ picking success rate after 5-shot adaptation to the YuMi robot with the different gripper sizes. These results validate that our approach can scale to a real robot with image observations.

\section{Conclusion}
The primarily goal of this study was to answer the following question: \textit{is it possible to obtain a meta-learner that captures the common structure shared across different robotic platforms such that a policy can be adapted to a novel platform  in an unambiguous and data-efficient manner?} We approached this problem by introducing a MAML-based probabilistic meta-learning framework where the meta-tasks are represented by a low-dimensional meta-task latent variable. We exploited the latent meta-task variable for modeling the meta-task uncertainties originating from the few-shot formulation of the policy adaptation problem. 
We studied the performance of our proposed framework on a reaching task performed in simulation on 400 robots generated with various physical parameters of the existing robotic platforms. Our method obtained superior results compared to the benchmarking MAML, VERSA and AVI frameworks. The adaptation of the policy was particularly successful to the robotic platforms whose variations of the physical parameters were present in the meta-train dataset. However, the performance of all considered frameworks dropped when the adaptation was done on a novel robotic platform without a similar representative in the meta-train dataset. We leave such out-of-distribution cases as well as evaluation on more complex visuomotor scenarios using several more platforms as future work. Given the promising results in this study, we ultimately aim to apply our framework on a multi-robot multi-task scenario.

\section*{Acknowledgments}
This work was supported by Knut and Alice Wallenberg Foundation, the EU through the project EnTimeMent and the Swedish Foundation for Strategic Research through the COIN project.

\bibliographystyle{IEEEtran}
\balance
\bibliography{ref}

\end{document}